\pdfoutput=1
\documentclass[11pt]{article}

\usepackage[final]{acl}

\usepackage{times}
\usepackage{latexsym}

\usepackage[T1]{fontenc}
\usepackage[utf8]{inputenc}

\usepackage{microtype}

\usepackage{inconsolata}

\usepackage{graphicx}

\usepackage{amssymb}
\usepackage{pifont}% http://ctan.org/pkg/pifont

\usepackage{arydshln}

\usepackage{xcolor}
\definecolor{main-color}{rgb}{0.6627, 0.7176, 0.7764}
\definecolor{back-color}{rgb}{0.1686, 0.1686, 0.1686}
\definecolor{string-color}{rgb}{0.3333, 0.5254, 0.345}
\definecolor{key-color}{rgb}{0.8, 0.47, 0.196}

\usepackage{listings}
\lstdefinestyle{mystyle}
{
    basicstyle=\small,
    keywordstyle = {\color{key-color}},
    keywordstyle = [2]{\color{blue}},
    morekeywords = [2]{intent_1,intent_2,intent_3,intent_N},
    morekeywords = [2]{previous_utterance_1, previous_utterance_2,previous_utterance_3},
    morekeywords = [2]{utterance_to_classify},
}

\title{Intent Recognition and Out-of-Scope Detection using LLMs \\ in Multi-party Conversations}

\author{Galo Castillo-L\'{o}pez \quad Ga\"{e}l de Chalendar \quad Nasredine Semmar \\
        Universit\'{e} Paris-Saclay, CEA, List, Palaiseau, France \\ \{\texttt{galo-daniel.castillolopez, gael.de-chalendar, nasredine.semmar}\}\texttt{@cea.fr}  }

\begin{document}
\maketitle
\begin{abstract}
Intent recognition is a fundamental component in task-oriented dialogue systems (TODS). Determining user intents and detecting whether an intent is Out-of-Scope (OOS) is crucial for TODS to provide reliable responses. However, traditional TODS require large amount of annotated data. In this work we propose a hybrid approach to combine BERT and LLMs in zero and few-shot settings to recognize intents and detect OOS utterances. Our approach leverages LLMs generalization power and BERT's computational efficiency in such scenarios. We evaluate our method on multi-party conversation corpora and observe that sharing information from BERT outputs to LLMs leads to system performance improvement.

\end{abstract}

\section{Introduction}
\label{sec:introduction}
\let\thefootnote\relax\footnotetext{This paper has been accepted for publication at SIGDIAL 2025 and corresponds to the author's version of the work.}
Advances in dialogue systems have facilitated their employment to assist users on daily tasks in domains such as banking, health consulting, hospitality and others \cite{valizadeh2022ai,camilleri2023chatbot,casanueva2020efficient}. Task-oriented dialogue systems (TODS) in real-world applications must be able to both recognize user intents and detect Out-of-Scope (OOS) intents to generate reliable responses. Standard methods for intent recognition generally require large amounts of annotated data. However, annotations are scarce in some real-world applications, especially when new intents are introduced into systems. Large Language Models have been shown to be robust at classification tasks in zero and few-shot settings. Nevertheless, inferences from LLMs are computationally costly, thus their extensive use remains unpractical in some scenarios. Previous work have proposed hybrid approaches combining LLMs and smaller language models, by only routing uncertain inferences to LLMs at inference time \cite{arora-etal-2024-intent}. These approaches consist in processing queries in two steps: first through the computationally efficient model, and then through LLMs, if necessary. In doing so, overall computational costs are reduced without compromising prediction quality. However, these methods do not share information among models, and hence LLMs miss potential relevant information from the preceding step. 

\begin{figure}[t]
\centering
\includegraphics[width=0.95\linewidth]{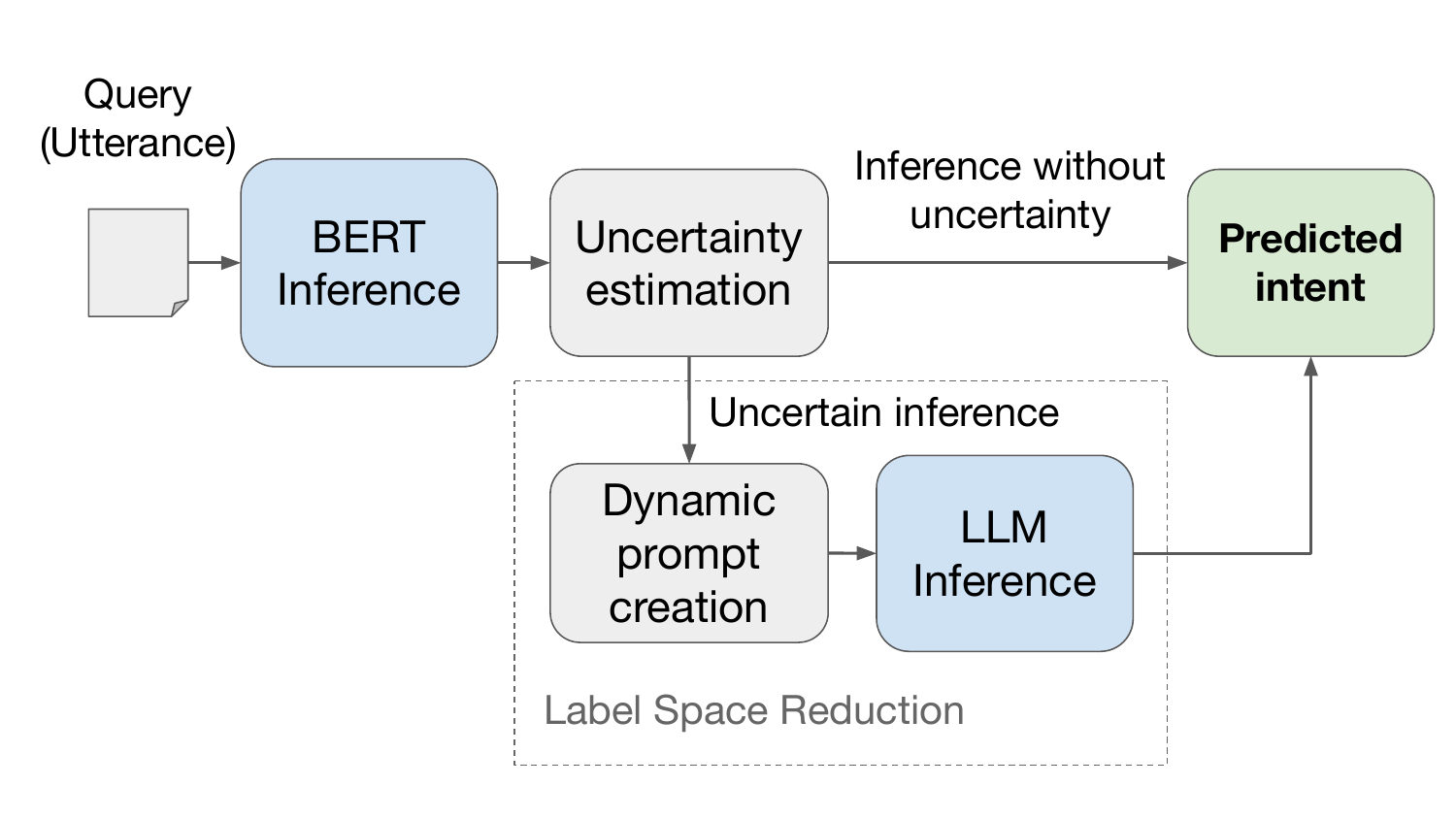}
\caption{Overview of our Label Space Reduction method.}
\label{fig:overview_lsr}
\end{figure}

In this work we propose a hybrid approach that combines small language models and LLMs for intent recognition and OOS detection in multi-party conversations, i.e. dialogues between three or more participants. We route inferences with high uncertainty from fine-tuned BERT models to LLMs, and use the information from the outputs of the fine-tuned models to dynamically generate the prompts at inference time. Such information is employed to reduce the label space on the classification task. Experiments in this study are conducted on three open source LLMs. Our work leverages the efficiency of (relatively) small models and the power of LLMs in zero-shot settings for intent classification and OOS detection. Figure \ref{fig:overview_lsr} illustrates our proposed method.

\section{Related Work}
\label{sec:related}
In recent years, intent detection methods have mainly consisted in fine-tuning small models \cite{larson-etal-2019-evaluation,arora-etal-2020-hint3,wang-etal-2023-app}. \citet{gautam-etal-2024-class} studied the use of class names to improve in-scope (IS) intent classification and OOS detection, using BERT and Spherical Variational Autoencoders \cite{davidson2018hyperspherical}. \citet{vishwanathan-etal-2022-multi} observed that fine-tuning sentence transformers presents largely better IS and OOS performance than traditional methods.

LLMs have gained attention in multiple NLP tasks, including intent recognition in dialogue systems \cite{lin-etal-2024-generate,wang-etal-2024-beyond,shin2024learning}. Findings are contradictory, as some works have found that LLMs outperform fine-tuned models \cite{addlesee2023multiparty} and others have shown the opposite \cite{zhang-etal-2024-new-approach}. To the best of our knowledge, the only study on intent recognition and OOS detection with a focus on efficiently using LLMs is proposed in \cite{arora-etal-2024-intent}. They propose a hybrid method that uses sentence transformers and LLMs, which reduced the performance gap to 2\% while reducing computing latency up to 50\%. However, their approach does not consider sharing information among models. Furthermore, we focus our work in multi-party conversations as such scenarios have been overlooked in previous work across most dialogue system tasks \cite{ganesh2023survey,castillo2025survey}.

\section{Experimental Procedure}
\label{sec:experiments}

\begin{table*}[ht]
\small
\centering
% $\heartsuit\varheart\diamondsuit\vardiamond\clubsuit\spadesuit$
\begin{tabular}{l | c c c | c c c}
% \begin{tabular}{l l c c c : c c c}
 \hline
 & \multicolumn{3}{c|}{In-scope} & \multicolumn{3}{c}{In-scope + Out-of-scope}\\
\cline{2-7}

Methods & ACC & WF1 & WP & ACC & F1-OOS & F1\\ 

\hline

ChatGPT\textsubscript{zero}\textsuperscript{$\spadesuit$}  & 35.27 & 37.10 & 48.22 & 27.68 & 21.21 & 28.34 \\

% \hdashline %\hdashline[0.5pt/5pt]

Mixtral 8$\times$7B\textsubscript{zero} & 31.87 & 32.17 & \textbf{51.35} & 31.46 & \textbf{38.66} & 26.97 \\

Llama-3 70B\textsubscript{zero} & 36.65 & 36.87 & 47.10 & 25.54 & 11.64 & 27.88 \\

DeepSeek-R1 70B\textsubscript{zero} & \textbf{41.22} & \textbf{43.47} & 49.99 & \textbf{35.79} & 35.06 & \textbf{35.14} \\

\hdashline %\hdashline[0.5pt/5pt]

MAG-BERT\textsubscript{ten}\textsuperscript{$\spadesuit$} & 9.82 & 11.58 & 13.34 & 34.58 & \textcolor{blue}{\textbf{50.57}} & 3.75 %\footnotemark
\\

ChatGPT\textsubscript{ten}\textsuperscript{$\spadesuit$} & 34.53 & 36.39 & 49.27 & 29.72 & 27.85 & 28.41 \\

BERT\textsubscript{ten} & 10.53 & 15.20 & 47.38 & \textcolor{blue}{\textbf{34.64}} & 49.35 & 16.68 \\

BERT\textsubscript{ten} $+$ Mixtral 8$\times$7B\textsubscript{zero} & 33.34 & 34.26 & 47.83 & 29.60 & 31.95 & 28.65 \\

BERT\textsubscript{ten} $+$ Llama-3 70B\textsubscript{zero} & 37.38 & 37.92 & 48.33 & 25.33 & 8.67 & 29.03 \\

BERT\textsubscript{ten} $+$ DeepSeek-R1 70B\textsubscript{zero} & \textbf{41.42} & \textbf{43.51} & \textbf{49.58} & 33.41 & 28.46 & \textbf{35.05} \\

\hdashline %\hdashline[0.5pt/5pt]

BERT\textsubscript{ten} $+$ Mixtral 8$\times$7B\textsubscript{zero} \textbf{\small{(LSR)}} & 35.02 & 35.76 & 47.03 & 28.61 & 25.98 & 29.96 \\

BERT\textsubscript{ten} $+$ Llama-3 70B\textsubscript{zero} \textbf{\small{(LSR)}} & 39.45 & 40.00 & 49.88 & 26.32 & 7.36 & 31.21 \\

BERT\textsubscript{ten} $+$ DeepSeek-R1 70B\textsubscript{zero} \textbf{\small{(LSR)}} & \textcolor{blue}{\textbf{41.66}} & \textcolor{blue}{\textbf{44.82}} & \textcolor{blue}{\textbf{52.65}} & \textbf{33.96} & \textbf{29.48} & \textcolor{blue}{\textbf{36.55}} \\

\hdashline %\hdashline[0.5pt/5pt]

Humans\textsubscript{ten}\textsuperscript{$\spadesuit$} & \textit{64.34} & \textit{67.82} & \textit{72.80} & \textit{60.43} & \textit{62.83} & \textit{57.83} \\

 \hline

\end{tabular}
% }
\caption{Results on the MIntRec 2.0 Corpus. Learning strategies include fine-tuning in \textit{ten}-shot as well as \textit{zero}-shot prompting. Results from \cite{zhang2024mintrec2} are denoted with $\spadesuit$. Our results implement a label space reduction approach \textbf{(LSR)} leveraging BERT probability outputs. IS evaluation metrics include accuracy (ACC), weighted F1 (WF1) and weighted precision (WP). IS+OOS settings are evaluated on accuracy (ACC), macro F1 (F1), and F1 score on the out-of-scope label (F1-OOS). Scores in \textbf{bold} highlight the best performing model per setting, and scores in \textcolor{blue}{\textbf{blue}} highlight the best performances overall.}
\label{table:results_MIntRec2.0}
\end{table*}

\begin{table*}[ht]
\small
\centering
% $\heartsuit\varheart\diamondsuit\vardiamond\clubsuit\spadesuit$
\begin{tabular}{l | c c c | c c c}
% \begin{tabular}{l l c c c : c c c}
 \hline
 & \multicolumn{3}{c|}{In-scope} & \multicolumn{3}{c}{In-scope + Out-of-scope}\\
\cline{2-7}

Methods & ACC & WF1 & WP & ACC & F1-OOS & F1\\ 

\hline

Mixtral 8$\times$7B\textsubscript{zero} & 65.41 & 67.38 & 83.64 & 56.25 & 17.65 & 51.02 \\

Llama-3 70B\textsubscript{zero} & 87.22 & 86.55 & 86.43 & 73.12 & 6.25 & 55.18 \\

DeepSeek-R1 70B\textsubscript{zero} & \textbf{89.47} & \textbf{91.17} & \textbf{93.65} & \textbf{78.75} & \textbf{35.90} & \textcolor{blue}{\textbf{75.17}} \\

\hdashline %\hdashline[0.5pt/5pt]

BERT\textsubscript{ten} & 50.38 & 58.92 & \textcolor{blue}{\textbf{96.89}} & 56.25 & \textbf{40.70} & 64.65 \\

BERT\textsubscript{ten} $+$ Mixtral 8$\times$7B\textsubscript{zero} & 72.93 & 74.65 & 88.28 & 61.88 & 13.33 & 63.53 \\

BERT\textsubscript{ten} $+$ Llama-3 70B\textsubscript{zero} & 90.23 & 90.23 & 90.98 & 75.00 & 0.0 & 62.98 \\

BERT\textsubscript{ten} $+$ DeepSeek-R1 70B\textsubscript{zero} & \textbf{90.98} & \textbf{92.30} & 94.30 & \textbf{79.38} & 33.33 & \textbf{73.15} \\

\hdashline %\hdashline[0.5pt/5pt]

BERT\textsubscript{ten} $+$ Mixtral 8$\times$7B\textsubscript{zero} \textbf{\small{(LSR)}} & 72.18 & 80.49 & \textbf{95.28} & 69.38 & 40.54 & 68.57 \\

BERT\textsubscript{ten} $+$ Llama-3 70B\textsubscript{zero} \textbf{\small{(LSR)}} & 89.47 & 90.73 & 92.43 & 75.62 & 12.12 & 67.21 \\

BERT\textsubscript{ten} $+$ DeepSeek-R1 70B\textsubscript{zero} \textbf{\small{(LSR)}} & \textcolor{blue}{\textbf{91.73}} & \textcolor{blue}{\textbf{92.70}} & 93.97 & \textcolor{blue}{\textbf{81.88}} & \textcolor{blue}{\textbf{45.00}} & \textbf{73.08} \\

 \hline

\end{tabular}
% }
\caption{Results on the MPGT Corpus. Learning strategies include fine-tuning in \textit{ten}-shot as well as \textit{zero}-shot prompting. Our results implement a label space reduction approach \textbf{(LSR)} leveraging BERT probability outputs. IS evaluation metrics include accuracy (ACC), weighted F1 (WF1) and weighted precision (WP). IS+OOS settings are evaluated on accuracy (ACC), macro F1 (F1), and F1 score on the out-of-scope label (F1-OOS). Scores in \textbf{bold} highlight the best performing model per setting, and scores in \textcolor{blue}{\textbf{blue}} highlight the best performances overall.}
\label{table:results_MPGT}
\end{table*}

Let $D=\{(u_i,y_i) \mid y_i \in \mathcal{Y}_A\}$ be a labeled dataset where $u_i$ denotes the $i_{th}$ utterance labeled with intent $y_i$, and $\mathcal{Y}_A=\{ 1, \dots, m, m+1\}$ denotes the set of $m$ in-scope intents plus the out-of-scope label. Our aim is to build a multiclass classification system that detects whether $u_i$ corresponds to an OOS intent from an unknown distribution or whether $u_i$ can be classified into any of $m$ possible in-scope intents.

\subsection{Datasets}
\label{sec:datasets}
We use two multi-party conversations corpora in this work. The first corpus is \textbf{MIntRec2.0}, a multimodal dataset of 15K multi-party dialogues from TV shows \cite{zhang2024mintrec2}. Modalities include audio, video and transcripts. In this study we are interested in systems working with text input data, thus we only use the text modality. The second dataset is \textbf{MPGT}, which is a collection of 29 multi-party dialogues between users and a receptionist robot in a hospital \cite{addlesee2023multiparty}. The MIntRec2.0 and MPGT datasets contain 30 and 8 in-scope intents, respectively, and both count with OOS utterances. Additional information about the datasets is detailed in Appendix \ref{sec:corpora_details}.

\subsection{Methods}
We evaluate four different approaches for intent recognition and OOS utterance detection: fine-tuned BERT; LLMs zero-shot classification; Uncertainty-based Query Routing combining BERT and LLMs, following the strategy proposed in \cite{arora-etal-2024-intent}; and our proposed Label Space Reduction method (LSR) using BERT inference outputs and LLMs. We detail such methods below.

\subsubsection{Small Language Model Fine-tuning}
\label{sec:bert_fine_tuning}
\paragraph{Fine-tuning.} We use the same pre-trained BERT language model as in \cite{zhang2024mintrec2}. Since having large number of examples per intent is challenging in real-world scenarios, especially when introducing new intents into systems, we conduct experiments on ten-shot settings following a similar approach to \cite{zhang2024mintrec2}. In addition, we concatenate each utterance with its 3 preceding utterances to enhance the model performance by introducing context information. A special turn-shift token \texttt{<ts>} is included between each pair of concatenated utterances to explicitly indicate change of turns in dialogues. Thus, each input example is a text sequence corresponding to an utterance with its concatenated context. We fine-tune the uncased version of BERT\textsubscript{BASE} \cite{devlin2019bert} for multiclass classification over the entire set of $m$ in-scope classes. Fine-tuning is performed over 5 different seeds to compute uncertainty scores from multiple runs at inference time. We detail the hyperparameter set we use on BERT-fine-tuning experiments on Appendix \ref{sec:app_bert}. At inference time, the predicted class is obtained by a majority voting strategy. Note that in this approach, the pre-trained model is fine-tuned without the OOS label, thus a OOS class detection strategy is needed.

\paragraph{Out-of-Scope Detection.} In order to detect OOS intents from our fine-tuned models, we quantify model uncertainty from the 5 outputs by computing the standard deviation of the softmax function applied on the logits in the last layer of the models (i.e. the probability estimates). Analysis on the validation sets showed that standard deviations $\sigma = 0.10$ and $\sigma = 0.12$ on the fine-tuned model probabilities provide good performance while maintaining a balance between OOS recall and IS macro F1-score, on the MIntRec2.0 and MPGT datasets, respectively.

\subsubsection{Large Language Models}
Large Language Models have been shown to excel at various classification tasks in zero-shot settings. We use three mid-sized instruct-tuned versions of open source LLMs: Mixtral8$\times$7B, LLaMA-3 70B, and DeepSeek-R1 70B (distilled). More details about the used LLMs can be found in Appendix \ref{sec:app_llms}. Our experiments on all LLMs are conducted on zero-shot prompting and use the same prompt template. The prompts we use describe the classification task; list the possible intents; define an OOS label; provide context from preceding utterances; define the expected output format; and include the utterance to classify. The prompt template is shown in Figure \ref{fig:prompt_template}. We investigate how LLMs alone perform in our classification task, as well as in combination with BERT, as described in further sections. In contrast to the OOS detection strategy used on our fine-tuned models, we do not need to add an additional step at inference time as our prompts already instructs either recognizing intents or detecting OOS samples. In other words, the OOS detections are directly obtained from the LLMs.

\subsubsection{Uncertainty-based Query Routing}
\label{sec:uncertainty}
Following the uncertainty-based query routing strategy proposed in \cite{arora-etal-2024-intent}, we combine BERT and LLMs by dispatching uncertain inferences made by BERT to LLMs. By doing so, only examples with high uncertainty are handled by LLMs, and costs due to the use of LLMs are reduced. We use the output probabilities by the 5 fine-tuned models and compute their standard deviation to quantify the uncertainty of the prediction, as explained in \ref{sec:bert_fine_tuning}. Prompts used in these experiments are the same as in the only-LLMs approach, where models are instructed to classify utterances into any of the IS intents or determine whether the utterance is OOS.

\subsubsection{Label Space Reduction}
We propose leveraging the outputs from the fine-tuned language models, and using such information to dynamically create prompts for LLM inference on routed queries. Our method extends the strategy described in Section \ref{sec:uncertainty}. Instead of including all labels on the LLM prompts, we consider the intents with the highest probabilities (i.e. estimates from the softmax function on the final layer logits) outputted by the fine-tuned models. The intent set selection is conducted as follows. For every routed utterance $u_i$, we retrieve subset $K_i$ of top-ranked intents whose cumulative sum of probabilities is at least $P$. The subset $K_i$ is the smallest subset of intents defined as $K_i = \{ y_1, y_2, \dots, y_k \} \subseteq \mathcal{Y}_S$ such that $\sum_{j=1}^{k} p_i(y_j) \geq P$, where $\mathcal{Y}_S$ is the full set of in-scope intents, $p_i(y_j)$ is the softmax probability of label $j$ for inference on $u_i$, and $P$ is a hyperparameter that controls the label space reduction (LSR). Lower values of $P$ result in higher space reduction. Therefore, the amount of intents included on the routed LLM inferences vary among examples. We found on the validation sets that $P=0.85$ achieves average hit rates slightly above 90\% on the intent subsets on both datasets, while reducing the label spaces on average by $\approx$80\% and $\approx$50\% on the MIntRec2.0 and MPGT corpora, respectively. This suggests that our approach retrieves pertinent labels after label filtering.

\subsection{Evaluation}
Method evaluations are performed in IS and IS+OOS scenarios. In-scope evaluation does not consider test examples belonging to the OOS label, whereas IS+OOS considers all labels including the OOS label. We follow previous work \cite{zhou2024token,zhang2024mintrec2,chen2024dual} and adopt three metrics for IS evaluation: Accuracy (ACC), Weighted F1 (WF1), and Weighted Precision (WP). Similarly, we use three commonly used metrics for IS+OOS evaluation: Accuracy (ACC) and F1-score (F1) over all classes, as well as F1-score over the OOS label (F1-OSS).

% \paragraph{Comparative Baselines.} We compare our proposed method with the results on few-shot learning approaches obtained in \cite{zhang2024mintrec2}. These approaches include: (1) MAG-BERT\textsubscript{ten}, a multimodal architecture trained on ten examples; (2) ChatGPT\textsubscript{zero} and ChatGPT\textsubscript{ten}, an LLM-based approach using zero and ten-shot in-context learning, respectively; and (3) Humans\textsubscript{ten}, an evaluation on humans instructed with ten examples. We also follow a comparable approach to the hybrid method proposed by \cite{arora-etal-2024-intent}, where no label space reduction is performed.

\section{Results}
\label{sec:results}
\begin{figure}[!htb]
\centering
\includegraphics[width=1.0\linewidth]{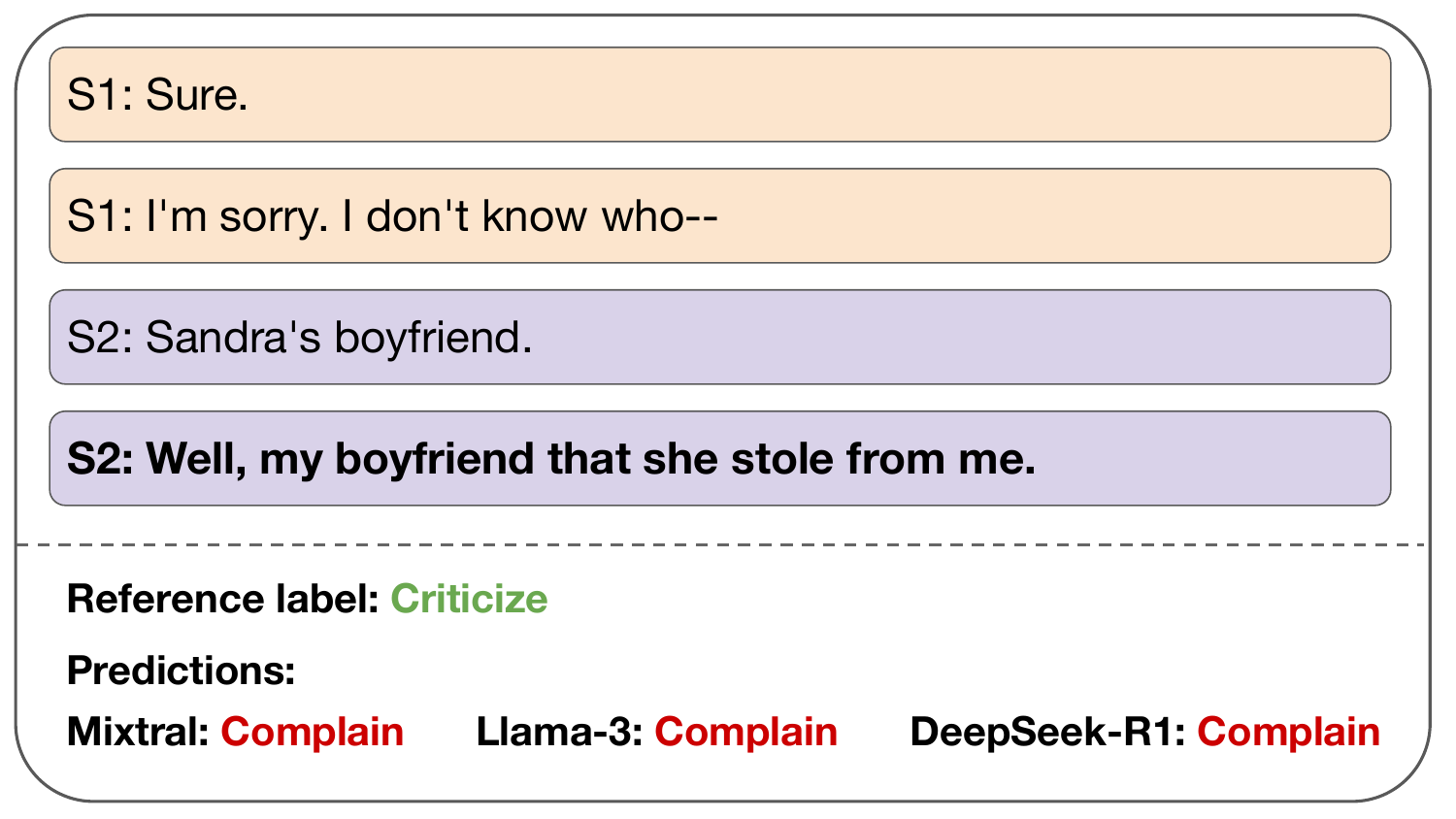}
\caption{Example of an instance difficult to classify by LLMs from the MIntRec2.0 corpus.}
\label{fig:example_error_1}
\end{figure}

Table \ref{table:results_MIntRec2.0} shows the results of our experiments on the MIntRec2.0 corpus. We observe that the best overall results in all in-scope performance metrics are obtained by our method on DeepSeek-R1. Reducing the label space results in an increase of $\thickapprox$3\% on the weighted precision. We also observe that when comparing the same BERT+LLM combinations, with and without label space reduction, better in-scope performance is obtained when reducing the label space in most metrics. BERT and MAG-BERT present the best overall performance on OOS evaluation. Nevertheless, their generalization on in-scope intents are the lowest compared to the other approaches. Additionally, as our method routes utterances with high uncertainty --i.e. potential OOS intents-- to LLMs, it is expected to see a decrease on the F1-OSS score (in particular, a decrease on the OOS recall). It is also noteworthy that the classification task is complex even for humans, according to the results reported by \citet{zhang2024mintrec2}. We believe that such complexity might be due to a high number of intents and the presence of overlapping intents on annotations. An example of a difficult instance to classify by LLMs is displayed in Figure \ref{fig:example_error_1}. We observe that similar to the example shown in Figure \ref{fig:example_error_1}, multiple other instances from the MIntRec2.0 corpus semantically overlap with more than one intent.

Results on the MPGT corpus are found in Table \ref{table:results_MPGT}, which show that DeepSeek-R1 with label space reduction obtains the best overall results in all metrics, except on in-scope WP and IS+OOS F1. In contrast to the results we observed on the MIntRec2.0 corpus, our method outperforms the fine-tuned BERT model alone on the F1-OOS score by $\thickapprox$5\%. In fact, all LLMs show to enhance their OOS detection when reducing the label space. We argue that LLMs struggle to detect OOS intents (more than smaller language models) when there is a higher number of IS intents, as suggested in \cite{wang-etal-2024-beyond}. In line with the results on Table \ref{table:results_MIntRec2.0}, Llama-3 is the worst OOS detector in all settings. An increase between 8\% and 12\% on the IS+OOS accuracy is observed when reducing the label space on Mixtral and DeepSeek-R1. We also conduct additional analysis on the impact of the label space reduction hyperparameter $P$ on the MPGT corpus in Appendix \ref{sec:p_hyperparam}.

\section{Computational Efficiency Analysis}
\label{sec:efficiency}
Table \ref{table:efficiency} shows a computational efficiency comparison between the use of BERT, LLMs, and the proposed label space reduction approach on the MPGT corpus. Note that our analysis considers all the 5 runs on the BERT inferences, which are performed to estimate inference uncertainty. We observe that our proposed method reduces the computational costs in more than 40\% when combining BERT with Llama-3 70B and Mixtral 8$\times$7B. To perform fair comparisons among methods and models, we employ the same computational resources on all inferences in this analysis.

\begin{table}[htb]
\small
\centering
\renewcommand{\arraystretch}{1.2}  % Provide more space between table rows, if you prefer
\begin{tabular}{lrr} \hline
 & Avg. & Latency \\ 
Methods & latency & ratio \\  \hline
Mixtral 8$\times$7B\textsubscript{zero} & 1.925 & \\
BERT\textsubscript{ten} & 0.065 & 0.034 \\
BERT\textsubscript{ten} $+$ Mixtral 8$\times$7B\textsubscript{zero} (LSR) & 1.100 & 0.571 \\

\hline
Llama-3 70B\textsubscript{zero} & 4.039 &  \\
BERT\textsubscript{ten} & 0.065 & 0.016 \\
BERT\textsubscript{ten} $+$ Llama-3 70B\textsubscript{zero} (LSR) & 2.236 & 0.553 \\

\hline\end{tabular}
\caption{Method efficiency comparison on the MPGT corpus. Comparison is based on average latency per inference (seconds) and the latency ratio with respect to the zero-shot LLM inference method (without label space reduction).}
\label{table:efficiency}
\end{table}

\section{Conclusions}
\label{sec:conclusions}
We investigated how (relatively) small language models such as BERT can be combined with LLMs in zero-shot scenarios to reduce computational costs on intent recognition tasks without compromising predictive quality. Our results on MPCs are in line with previous works in dyadic dialogues, suggesting that uncertainty-based routing lead to performance gains. Our work also demonstrates that sharing information among models such as probability estimates to reduce the label space outperforms methods without shared information. Future work may consider exploring other plausible label selection strategies. Additionally, other sources of information to be leveraged in LLM prompts from small models (i.e. BERT) may be investigated in future studies: the actual probability estimates, uncertainty patterns, model's internal representations, etc. Finally, although our experiments are conducted on multi-party corpora, our proposed method could also be applied on dyadic scenarios. We believe that our findings show promising directions towards robust and efficient intent recognition systems in real-world applications.

%\section*{Limitations}
% By default, the box containing the title and author names is set to the minimum of 5 cm. If you need more space, include the following in the preamble:
% \begin{quote}
% \begin{verbatim}
% \setlength\titlebox{<dim>}
% \end{verbatim}
% \end{quote}
% where \verb|<dim>| is replaced with a length. Do not set this length smaller than 5 cm.

% \input{content/limitations}
\section*{Ethical Considerations}
In developing our hybrid approach for intent detection using BERT and LLMs, we considered several ethical implications to ensure responsible practices. Despite the use LLMs, which are capable of generating potential unsafe content, they are solely employed as text classifiers into sets of defined classes. Therefore, the risk of misuse or producing harmful content available for end users is minimized. However, it is important for any implementation of the proposed methods to be aware of potential biases inherent in those models. In addition, all our experiments use publicly available corpora, which have been curated prior to our work to prevent malicious actions. Overall, the contributions presented in this study are designed for constructive and ethical use, with no direct association with harmful social consequences.

\section*{Acknowledgments}
We warmly thank our anonymous reviewers for their time and valuable feedback. This publication was made possible by the use of the FactoryIA supercomputer, financially supported by the Ile-de-France Regional Council. This work has been partially funded by the EU project CORTEX2 (under grant agreement: N° 101070192).

% Bibliography entries for the entire Anthology, followed by custom entries
%\bibliography{anthology,custom}
% Custom bibliography entries only
\bibliography{acl_latex}

\begin{thebibliography}{25}
\providecommand{\natexlab}[1]{#1}

\bibitem[{Addlesee et~al.(2023)Addlesee, Siei{\'n}ska, Gunson, Garcia, Dondrup,
  and Lemon}]{addlesee2023multiparty}
Angus Addlesee, Weronika Siei{\'n}ska, Nancie Gunson, Daniel~Hern{\'a}ndez
  Garcia, Christian Dondrup, and Oliver Lemon. 2023.
\newblock Multi-party goal tracking with llms: Comparing pre-training,
  fine-tuning, and prompt engineering.
\newblock In \emph{Proceedings of the 24th Annual Meeting of the Special
  Interest Group on Discourse and Dialogue}.

\bibitem[{Arora et~al.(2020)Arora, Jain, Chaturvedi, and
  Modi}]{arora-etal-2020-hint3}
Gaurav Arora, Chirag Jain, Manas Chaturvedi, and Krupal Modi. 2020.
\newblock \href {https://doi.org/10.18653/v1/2020.insights-1.16} {{HINT}3:
  Raising the bar for intent detection in the wild}.
\newblock In \emph{Proceedings of the First Workshop on Insights from Negative
  Results in NLP}, pages 100--105, Online. Association for Computational
  Linguistics.

\bibitem[{Arora et~al.(2024)Arora, Jain, and Merugu}]{arora-etal-2024-intent}
Gaurav Arora, Shreya Jain, and Srujana Merugu. 2024.
\newblock \href {https://doi.org/10.18653/v1/2024.emnlp-industry.114} {Intent
  detection in the age of {LLM}s}.
\newblock In \emph{Proceedings of the 2024 Conference on Empirical Methods in
  Natural Language Processing: Industry Track}, pages 1559--1570, Miami,
  Florida, US. Association for Computational Linguistics.

\bibitem[{Camilleri and Troise(2023)}]{camilleri2023chatbot}
Mark~Anthony Camilleri and Ciro Troise. 2023.
\newblock Chatbot recommender systems in tourism: A systematic review and a
  benefit-cost analysis.
\newblock In \emph{Proceedings of the 2023 8th international conference on
  machine learning technologies}, pages 151--156.

\bibitem[{Casanueva et~al.(2020)Casanueva, Tem{\v{c}}inas, Gerz, Henderson, and
  Vuli{\'c}}]{casanueva2020efficient}
I{\~n}igo Casanueva, Tadas Tem{\v{c}}inas, Daniela Gerz, Matthew Henderson, and
  Ivan Vuli{\'c}. 2020.
\newblock Efficient intent detection with dual sentence encoders.
\newblock \emph{arXiv preprint arXiv:2003.04807}.

\bibitem[{Castillo-L{\'o}pez et~al.(2025)Castillo-L{\'o}pez, de~Chalendar, and
  Semmar}]{castillo2025survey}
Galo Castillo-L{\'o}pez, Ga{\"e}l de~Chalendar, and Nasredine Semmar. 2025.
\newblock A survey of recent advances on turn-taking modeling in spoken
  dialogue systems.
\newblock In \emph{Proceedings of the 15th International Workshop on Spoken
  Dialogue Systems Technology}, pages 254--271.

\bibitem[{Chen et~al.(2024)Chen, Zhu, Zhuang, Huang, and Zou}]{chen2024dual}
Zhanpeng Chen, Zhihong Zhu, Xianwei Zhuang, Zhiqi Huang, and Yuexian Zou. 2024.
\newblock Dual-oriented disentangled network with counterfactual intervention
  for multimodal intent detection.
\newblock In \emph{Proceedings of the 2024 Conference on Empirical Methods in
  Natural Language Processing}, pages 17554--17567.

\bibitem[{Davidson et~al.(2018)Davidson, Falorsi, De~Cao, Kipf, and
  Tomczak}]{davidson2018hyperspherical}
Tim~R Davidson, Luca Falorsi, Nicola De~Cao, Thomas Kipf, and Jakub~M Tomczak.
  2018.
\newblock Hyperspherical variational auto-encoders.
\newblock \emph{arXiv preprint arXiv:1804.00891}.

\bibitem[{DeepSeek-AI(2025)}]{deepseekai2025deepseekr1incentivizingreasoningcapability}
DeepSeek-AI. 2025.
\newblock \href {https://arxiv.org/abs/2501.12948} {Deepseek-r1: Incentivizing
  reasoning capability in llms via reinforcement learning}.
\newblock \emph{Preprint}, arXiv:2501.12948.

\bibitem[{Devlin et~al.(2019)Devlin, Chang, Lee, and
  Toutanova}]{devlin2019bert}
Jacob Devlin, Ming-Wei Chang, Kenton Lee, and Kristina Toutanova. 2019.
\newblock Bert: Pre-training of deep bidirectional transformers for language
  understanding.
\newblock In \emph{Proceedings of the 2019 conference of the North American
  chapter of the association for computational linguistics: human language
  technologies, volume 1 (long and short papers)}, pages 4171--4186.

\bibitem[{Ganesh et~al.(2023)Ganesh, Palmer, and von~der
  Wense}]{ganesh2023survey}
Ananya Ganesh, Martha Palmer, and Katharina von~der Wense. 2023.
\newblock A survey of challenges and methods in the computational modeling of
  multi-party dialog.
\newblock In \emph{Proceedings of the 5th Workshop on NLP for Conversational AI
  (NLP4ConvAI 2023)}, pages 140--154.

\bibitem[{Gautam et~al.(2024)Gautam, Parameswaran, Kane, Fang, Ramasamy,
  Sundaram, Sahu, and Li}]{gautam-etal-2024-class}
Chandan Gautam, Sethupathy Parameswaran, Aditya Kane, Yuan Fang, Savitha
  Ramasamy, Suresh Sundaram, Sunil~Kumar Sahu, and Xiaoli Li. 2024.
\newblock \href {https://doi.org/10.18653/v1/2024.findings-emnlp.531} {Class
  name guided out-of-scope intent classification}.
\newblock In \emph{Findings of the Association for Computational Linguistics:
  EMNLP 2024}, pages 9100--9112, Miami, Florida, USA. Association for
  Computational Linguistics.

\bibitem[{Grattafiori et~al.(2024)Grattafiori, Dubey, Jauhri, Pandey, Kadian,
  Al-Dahle, Letman, Mathur, Schelten, Vaughan et~al.}]{grattafiori2024llama}
Aaron Grattafiori, Abhimanyu Dubey, Abhinav Jauhri, Abhinav Pandey, Abhishek
  Kadian, Ahmad Al-Dahle, Aiesha Letman, Akhil Mathur, Alan Schelten, Alex
  Vaughan, and 1 others. 2024.
\newblock The llama 3 herd of models.
\newblock \emph{arXiv preprint arXiv:2407.21783}.

\bibitem[{Jiang et~al.(2024)Jiang, Sablayrolles, Roux, Mensch, Savary, Bamford,
  Chaplot, Casas, Hanna, Bressand et~al.}]{jiang2024mixtral}
Albert~Q Jiang, Alexandre Sablayrolles, Antoine Roux, Arthur Mensch, Blanche
  Savary, Chris Bamford, Devendra~Singh Chaplot, Diego de~las Casas, Emma~Bou
  Hanna, Florian Bressand, and 1 others. 2024.
\newblock Mixtral of experts.
\newblock \emph{arXiv preprint arXiv:2401.04088}.

\bibitem[{Larson et~al.(2019)Larson, Mahendran, Peper, Clarke, Lee, Hill,
  Kummerfeld, Leach, Laurenzano, Tang, and Mars}]{larson-etal-2019-evaluation}
Stefan Larson, Anish Mahendran, Joseph~J. Peper, Christopher Clarke, Andrew
  Lee, Parker Hill, Jonathan~K. Kummerfeld, Kevin Leach, Michael~A. Laurenzano,
  Lingjia Tang, and Jason Mars. 2019.
\newblock \href {https://doi.org/10.18653/v1/D19-1131} {An evaluation dataset
  for intent classification and out-of-scope prediction}.
\newblock In \emph{Proceedings of the 2019 Conference on Empirical Methods in
  Natural Language Processing and the 9th International Joint Conference on
  Natural Language Processing (EMNLP-IJCNLP)}, pages 1311--1316, Hong Kong,
  China. Association for Computational Linguistics.

\bibitem[{Lin et~al.(2024)Lin, Hasibi, and Verberne}]{lin-etal-2024-generate}
I-Fan Lin, Faegheh Hasibi, and Suzan Verberne. 2024.
\newblock \href {https://doi.org/10.18653/v1/2024.findings-emnlp.768} {Generate
  then refine: Data augmentation for zero-shot intent detection}.
\newblock In \emph{Findings of the Association for Computational Linguistics:
  EMNLP 2024}, pages 13138--13146, Miami, Florida, USA. Association for
  Computational Linguistics.

\bibitem[{Shin et~al.(2024)Shin, Ahn, Won, and Choi}]{shin2024learning}
Joongbo Shin, Youbin Ahn, Seungpil Won, and Stanley~Jungkyu Choi. 2024.
\newblock Learning to adapt large language models to one-shot in-context intent
  classification on unseen domains.
\newblock In \emph{Proceedings of the 1st Workshop on Customizable NLP:
  Progress and Challenges in Customizing NLP for a Domain, Application, Group,
  or Individual (CustomNLP4U)}, pages 182--197.

\bibitem[{Valizadeh and Parde(2022)}]{valizadeh2022ai}
Mina Valizadeh and Natalie Parde. 2022.
\newblock The ai doctor is in: A survey of task-oriented dialogue systems for
  healthcare applications.
\newblock In \emph{Proceedings of the 60th Annual Meeting of the Association
  for Computational Linguistics (Volume 1: Long Papers)}, pages 6638--6660.

\bibitem[{Vishwanathan et~al.(2022)Vishwanathan, Warrier, Vadakkekara~Suresh,
  and Kandpal}]{vishwanathan-etal-2022-multi}
Asha Vishwanathan, Rajeev Warrier, Gautham Vadakkekara~Suresh, and
  Chandra~Shekhar Kandpal. 2022.
\newblock \href {https://doi.org/10.18653/v1/2022.emnlp-industry.19}
  {Multi-tenant optimization for few-shot task-oriented {FAQ} retrieval}.
\newblock In \emph{Proceedings of the 2022 Conference on Empirical Methods in
  Natural Language Processing: Industry Track}, pages 188--197, Abu Dhabi, UAE.
  Association for Computational Linguistics.

\bibitem[{Wang et~al.(2023)Wang, He, Mou, Song, Wu, Wang, Xian, Cai, and
  Xu}]{wang-etal-2023-app}
Pei Wang, Keqing He, Yutao Mou, Xiaoshuai Song, Yanan Wu, Jingang Wang, Yunsen
  Xian, Xunliang Cai, and Weiran Xu. 2023.
\newblock \href {https://doi.org/10.18653/v1/2023.findings-emnlp.258} {{APP}:
  Adaptive prototypical pseudo-labeling for few-shot {OOD} detection}.
\newblock In \emph{Findings of the Association for Computational Linguistics:
  EMNLP 2023}, pages 3926--3939, Singapore. Association for Computational
  Linguistics.

\bibitem[{Wang et~al.(2024)Wang, He, Wang, Song, Mou, Wang, Xian, Cai, and
  Xu}]{wang-etal-2024-beyond}
Pei Wang, Keqing He, Yejie Wang, Xiaoshuai Song, Yutao Mou, Jingang Wang,
  Yunsen Xian, Xunliang Cai, and Weiran Xu. 2024.
\newblock \href {https://aclanthology.org/2024.lrec-main.210/} {Beyond the
  known: Investigating {LLM}s performance on out-of-domain intent detection}.
\newblock In \emph{Proceedings of the 2024 Joint International Conference on
  Computational Linguistics, Language Resources and Evaluation (LREC-COLING
  2024)}, pages 2354--2364, Torino, Italia. ELRA and ICCL.

\bibitem[{Wolf et~al.(2020)Wolf, Debut, Sanh, Chaumond, Delangue, Moi, Cistac,
  Rault, Louf, Funtowicz et~al.}]{wolf2020transformers}
Thomas Wolf, Lysandre Debut, Victor Sanh, Julien Chaumond, Clement Delangue,
  Anthony Moi, Pierric Cistac, Tim Rault, R{\'e}mi Louf, Morgan Funtowicz, and
  1 others. 2020.
\newblock Transformers: State-of-the-art natural language processing.
\newblock In \emph{Proceedings of the 2020 conference on empirical methods in
  natural language processing: system demonstrations}, pages 38--45.

\bibitem[{Zhang et~al.(2024{\natexlab{a}})Zhang, Wang, Xu, Zhou, Gao, Su, Li,
  Chen et~al.}]{zhang2024mintrec2}
Hanlei Zhang, Xin Wang, Hua Xu, Qianrui Zhou, Kai Gao, Jianhua Su, Wenrui Li,
  Yanting Chen, and 1 others. 2024{\natexlab{a}}.
\newblock Mintrec2. 0: A large-scale benchmark dataset for multimodal intent
  recognition and out-of-scope detection in conversations.
\newblock \emph{arXiv preprint arXiv:2403.10943}.

\bibitem[{Zhang et~al.(2024{\natexlab{b}})Zhang, Norouzian, Mohan, and
  Ducatelle}]{zhang-etal-2024-new-approach}
Tianyi Zhang, Atta Norouzian, Aanchan Mohan, and Frederick Ducatelle.
  2024{\natexlab{b}}.
\newblock \href {https://doi.org/10.18653/v1/2024.emnlp-industry.68} {A new
  approach for fine-tuning sentence transformers for intent classification and
  out-of-scope detection tasks}.
\newblock In \emph{Proceedings of the 2024 Conference on Empirical Methods in
  Natural Language Processing: Industry Track}, pages 910--919, Miami, Florida,
  US. Association for Computational Linguistics.

\bibitem[{Zhou et~al.(2024)Zhou, Xu, Li, Zhang, Zhang, Wang, and
  Gao}]{zhou2024token}
Qianrui Zhou, Hua Xu, Hao Li, Hanlei Zhang, Xiaohan Zhang, Yifan Wang, and Kai
  Gao. 2024.
\newblock \href {https://doi.org/10.1609/aaai.v38i15.29656} {Token-level
  contrastive learning with modality-aware prompting for multimodal intent
  recognition}.
\newblock In \emph{Proceedings of the {AAAI} Conference on Artificial
  Intelligence}, volume~38, pages 17114--17122.
\newblock Number: 15.

\end{thebibliography}
\appendix
\section{Model Information}
\label{sec:app_model_info}
In this appendix we provide model implementation details of our experiments.

\subsection{BERT Implementation Details}
\label{sec:app_bert}
All BERT fine-tuning runs were conducted on a single NVIDIA A100 GPU of 40GB. The average execution time for all fine-tuning experiments was less than 30 minutes to complete. We used the \texttt{BertForSequenceClassification} class from Hugging Face’s Transformers library \cite{wolf2020transformers} for sequence classification tasks. BERT\textsubscript{BASE} uncased is used in all the experiments. Table \ref{table:hyperparameters} shows the hyperparameter configuration we employ.

\begin{table}[!ht]
\centering
% \footnotesize
\begin{tabular}{lrr} \hline
hyperparameter & value\\ \hline
eval\_monitor & macro F1-score \\
train\_batch\_size & 16 \\
eval\_batch\_size & 16 \\
test\_batch\_size & 16 \\
wait\_patience & 3 \\
num\_train\_epochs & 40 \\
warmup\_proportion & 0.1 \\
lr & 1e-5 \\
\hline\end{tabular}
\caption{Set of hyperparameters used on BERT fine-tuning experiments.}
\label{table:hyperparameters}
\end{table}

\subsection{Large Language Models}
\label{sec:app_llms}
Our experiments on LLMs use mid-sized instruct versions of  models. Specifically, we use \texttt{Mixtral-8x7B-Instruct-v0.1} \cite{jiang2024mixtral}, \texttt{Meta-Llama-3-70B-Instruct} \cite{grattafiori2024llama}, and \texttt{DeepSeek-R1-Distill-Llama-70B} \cite{deepseekai2025deepseekr1incentivizingreasoningcapability}.

\subsection{Prompt Template}
\label{sec:app_prompt_template}
Figure \ref{fig:prompt_template} shows the prompt template we use on all LLM experiments.

% \begin{figure*}[t]
% \begin{lstlisting}[style=mystyle]
% **Task description**
% You are an out-of-domain intent detector, and your task is to detect whether the intent of the last utterance belong to the intents supported by the system, from dialogues of multiple participants. If they do, return the corresponding intent label, otherwise return UNK.

% **Authorized categories**
% The supported intents are:
% Explain, Emphasize, Agree, Doubt, Taunt, Joke, Inform, Complain, Flaunt

% **Out-of-domain label**
% - UNK

% **Previous utterances in the dialogue**
% You have the following utterance history from multiple participants to understand the context of the dialogue. Each utterance is on a line and starts by "-":
% - uh-huh.
% - uh, okay.
% - just that... you know, sometimes people flirt with people to get stuff.

% **Expected output format**
% Your response should only be a JSON object with the following structure:
% {"intent": "intent_label"}
% Do not write anything else.

% **Task**
% The utterance to classify is shown below:
% yeah, well, that's not what's happening here, so...

% Result:
% \end{lstlisting}
% \caption{Prompt design used with all LLMs in this work.}
% \end{figure*}[t]

\begin{figure*}[!htb]
\begin{lstlisting}[style=mystyle]
**Task description**
You are an out-of-domain intent detector, and your task is to detect whether the intent of the last utterance belongs to the intents supported by the system, from dialogues of multiple participants. If they do, return the corresponding intent label, otherwise return UNK.

**Authorized categories**
The supported intents are:
intent_1, intent_2, intent_3, ... intent_N

**Out-of-domain label**
- UNK

**Previous utterances in the dialogue**
You have the following utterance history from multiple participants to understand the context of the dialogue. Each utterance is on a line and starts by "-":
- previous_utterance_1
- previous_utterance_2
- previous_utterance_3

**Expected output format**
Your response should only be a JSON object with the following structure:
{"intent": "intent_label"}
Do not write anything else.

**Task**
The utterance to classify is shown below:
utterance_to_classify

Result:
\end{lstlisting}
\caption{Prompt template used on all LLM experiments. Highlighted text in \textcolor{blue}{blue} varies among dataset examples.}
\label{fig:prompt_template}
\end{figure*}

\section{Corpora Details}
\label{sec:corpora_details}
\subsection{MPGT Annotations}
\begin{figure}[!htb]
\centering
\includegraphics[width=\linewidth]{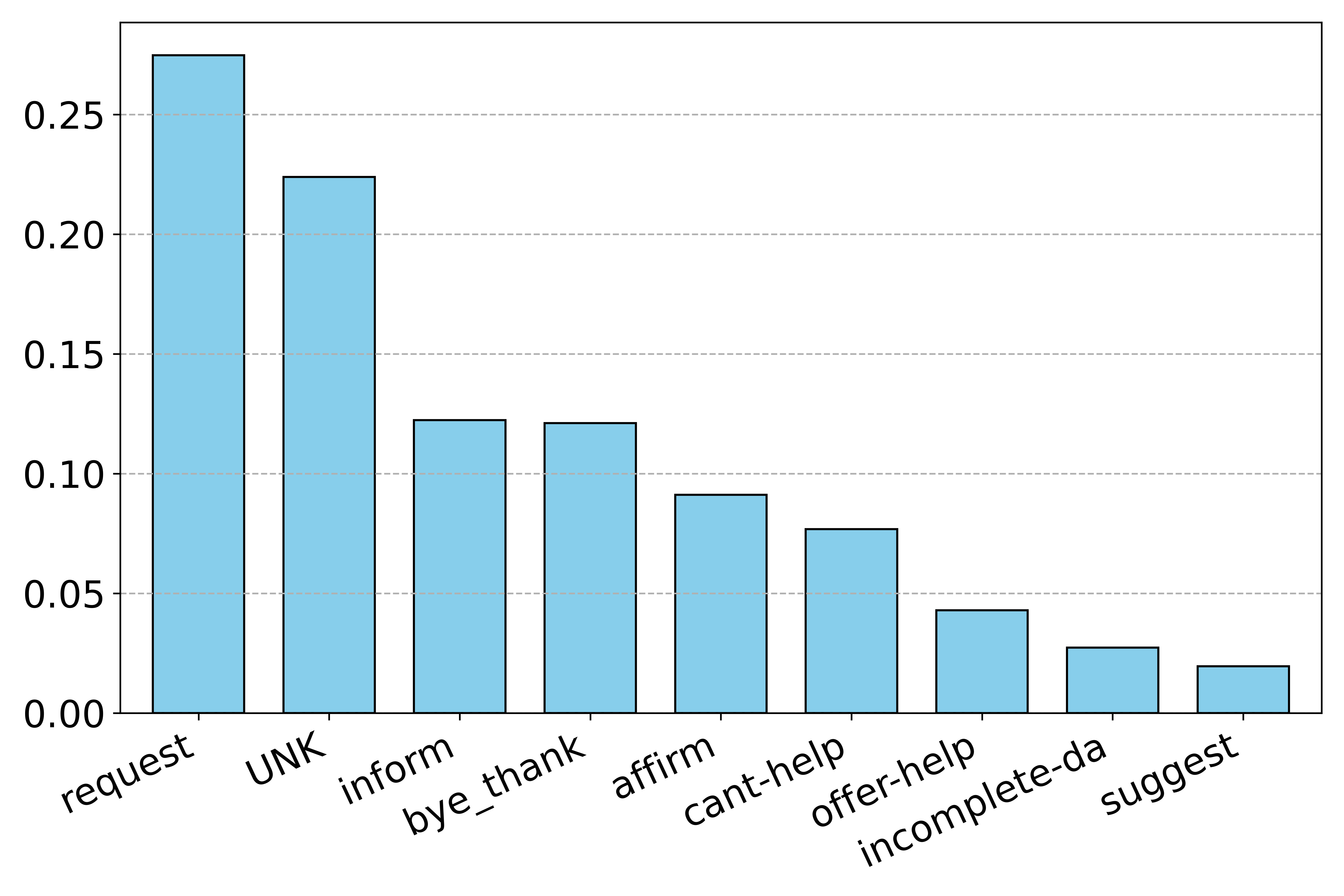}
\caption{Distribution of the intents in the adapted multi-class version of the MPGT corpus, including the OOS label (UNK).}
\label{fig:label_distribution_MPGT}
\end{figure}

In this work, we assume that every utterance corresponds to a single intent, either in-scope or out-of-scope. Thus, the intent recognition task can be defined as a multi-class classification problem. However, the MPGT corpus is built under the assumption that an utterance might belong to none, one, or many intents, i.e. multi-label classification. Hence, we adapted the corpus for multi-class intent classification through manual data curation and multiple strategies. These strategies consisted in combining co-occurring intents, grouping original labels and co-occurring combinations, and assigning the OOS label to rare/irrelevant intents. Figure \ref{fig:label_distribution_MPGT} shows the final label distribution after our adaptation. Our adapted multi-class version of the MPGT corpus for intent recognition is made available online\footnote{https://github.com/gaalocastillo/mpgt\_multiclass}.

% Upon acceptance for publication, we will release the adapted multi-class (intent recognition) version of the MPGT corpus to the community.

\subsection{Dataset Statistics}
\label{sec:dataset_info}
Table \ref{table:dataset_stats} shows statistics of the datasets we use in this work.
\begin{table}[htb]
\footnotesize
\centering
\renewcommand{\arraystretch}{1.5}  % Provide more space between table rows, if you prefer
\begin{tabular}{lrrrrr} \hline
% &\multicolumn{3}{c}{MIntRec2.0}&\multicolumn{3}{c}{MPGT} \\ \hline
% & \#dial & \#utt & \#utt & \%OOS \\ \hline
 & \#dial. & \#utt. & \#utt. fs. & \#int. & \%OOS \\ \hline
MintRec2.0 & 1.2K & 15K & 211 & 30 & 38\% \\
MPGT & 29 & 768 & 80 & 8 & 22\%  \\

\hline\end{tabular}
\caption{Dataset statistics: number of dialogues (\# dial.), number of utterances (\# utt.), number of utterances used on few-shot fine-tuning (\# utt. fs.), number of intent categories (\# int.), and proportion of OOS utterances (\%OOS).}
\label{table:dataset_stats}
\end{table}

\subsection{Subset Splits}
\label{sec:dataset_splits}
Table \ref{table:overview_splits} describes the subset splits we use for training, development and test. Note that our few-shot fine-tuning on BERT does not use all the training sets but only the selected few-shot utterances detailed in Table  \ref{table:dataset_stats} in Appendix \ref{sec:dataset_info}.
% \begin{table*}[!ht]
\begin{table}[htb]
\footnotesize
\centering
\renewcommand{\arraystretch}{1.5}  % Provide more space between table rows, if you prefer
\begin{tabular}{lrrrrrr} \hline
&\multicolumn{3}{c}{MIntRec2.0}&\multicolumn{3}{c}{MPGT} \\ \hline
 & train & dev & test & train & dev & test \\ \hline
\#dial. & 871 & 125 & 249 & 20 & 4 & 5 \\
\#utt. & 9.9K & 1.8K & 3.2K & 517 & 91 & 160 \\

\hline\end{tabular}
\caption{Number of dialogues (\# dial.) and utterances (\# utt.) per subset split.}
\label{table:overview_splits}
\end{table}

\section{Hyperparameter P}
\label{sec:p_hyperparam}
\begin{figure}[!htb]
\centering
\includegraphics[width=\linewidth]{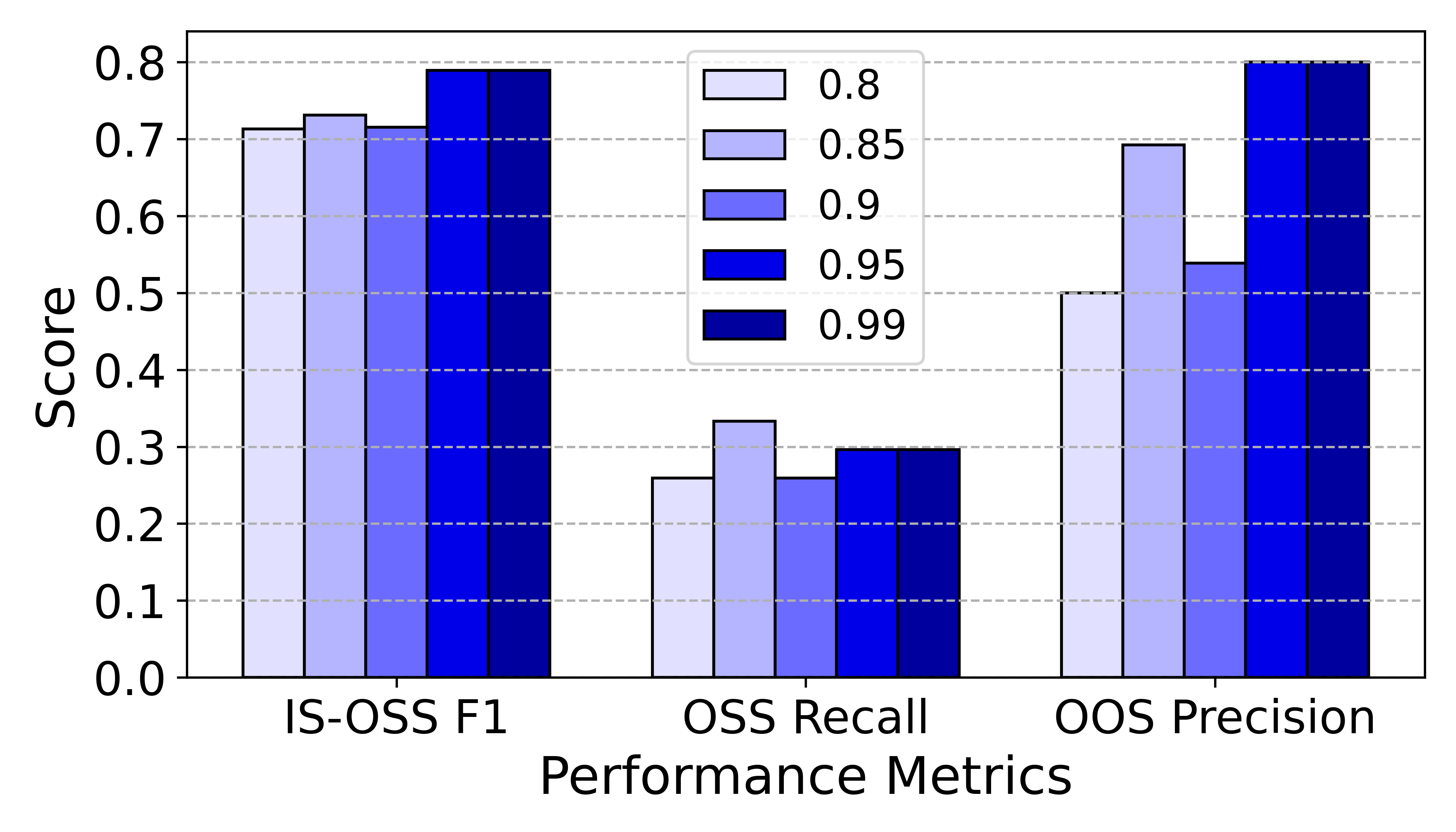}
\caption{Performance metrics at distinct values of the hyperparameter $P$ of our proposed method on the MPGT Corpus.}
\label{fig:hyperparam_p}
\end{figure}

Our proposed method relies on the hyperparameter $P$, which controls the label space reduction. Lower values of $P$ result in higher reduction, therefore less intents included in the LLM prompts. The main results of this paper, presented in Tables \ref{table:results_MIntRec2.0} and \ref{table:results_MPGT}, consider $P = 0.85$. We developed additional analysis on distinct values of P on the MPGT corpus and the BERT+DeepSeek method. Figure \ref{fig:hyperparam_p} suggests that low label space reduction ($P = 0.95$ and $P = 0.99$) presents better OOS precision and IS-OOS F1-score. Nevertheless, such improvement occurs at cost of missing OOS examples, as a decrease on the OOS recall is observed.

% This is an appendix.

\end{document}